\documentclass[11pt, a4paper,  notitlepage]{extarticle}

\usepackage[left=3.5cm, right=3.5cm, top=3cm, bottom=3cm, bindingoffset=0cm]{geometry}
\usepackage[T2A]{fontenc}
\usepackage[utf8]{inputenc}
\usepackage{graphicx}
\usepackage{mathtools}
\usepackage[affil-it]{authblk}
\usepackage{url}
\usepackage{amsmath}
\usepackage{amssymb}
\usepackage{multicol}
\usepackage{float}
\usepackage[section]{placeins}
\title{Methods of Informational Trends Analytics and Fake News Detection on Twitter}
\author{Bohdan M.  Pavlyshenko \\  \small{Ivan Franko National University of Lviv,  Ukraine \\  b.pavlyshenko@gmail.com,  www.linkedin.com/in/bpavlyshenko/ }}
\begin{document}
\maketitle
\sloppy
\begin{abstract}
In the paper, different approaches for the analysis of news trends on Twitter has been considered. For the analysis and case study,  informational trends on Twitter caused by Russian invasion of Ukraine in 2022 year have been studied.  A deep learning approach for fake news detection has been analyzed. The use of the theory of frequent  itemsets and association rules,   graph theory for news trends analytics have been considered.  

\end{abstract}

\section{Introduction}
News have an essential impact in many areas of society, politics and business. That is why one  can see a lot of attempts to produce manipulative and fake  news to get a specified response  in the society.  One of  horrible world events  is Russian invasion of Ukraine  on February, 24 2022.  It causes a large informational news flow on social networks, including producing manipulative and fake news to shape a specified explanation and justification of invasion. 
One of the goals claimed by Russia  was the  'denazification' of Ukraine. One of the allegations of Russian propaganda was  that Ukraine was developing the biological weapon in special laboratories. 

Tweets, the messages of Twitter microblogs,  have high density of semantically important keywords. It makes it possible to get semantically important information from tweets and generate the features of predictive models for the decision-making support. Different studies of Twitter are considered in the papers~\cite{java2007we,kwak2010twitter,pak2010twitter, cha2010measuring,benevenuto2009characterizing,
bollen2011twitter,asur2010predicting,shamma2010tweetgeist, 
wang2020novel, balakrishnan2020improving, grinberg2019fake,ajao2018fake,helmstetter2018weakly}. 
In~\cite{pavlyshenko2019forecasting,pavlyshenko2019cantwitter, pavlyshenko2021forming}, we study different approaches for the analysis of messages on Twitter, as well as the use of  tweet features for forecasting different kinds of events. 

In this paper, we consider the methods for the analysis of Twitter trends and for detecting fake news. 
As fake news,  we will consider the news information which is not true as well as the information which  can contain  real facts, but with incorrectly specified accents, and the focuses that lead to distorted conclusion and incorrect understanding of underlying processes.  For our analysis, we considered informational trends caused by Russian invasion of Ukraine in 2022. 
In the study, we also consider the possible impact of informational trends on different companies working in Russia during this conflict. 
\section{News Trends on Twitter}
\begin{figure}
\center
 \includegraphics[width=1\linewidth]{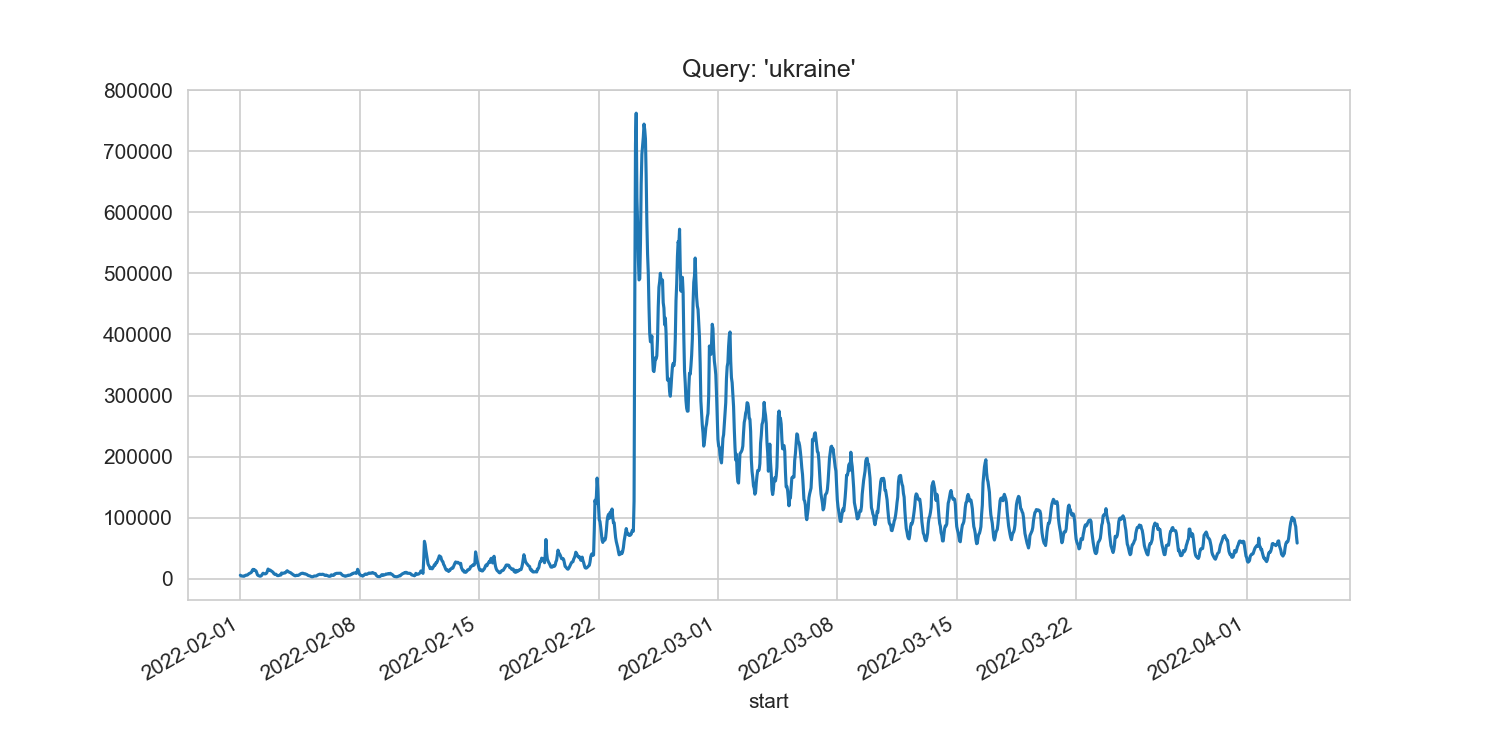}
 \caption{Time series of tweets for the query 'ukraine'}
 \label{ts1}
 \end{figure}

\begin{figure}
\center
 \includegraphics[width=1\linewidth]{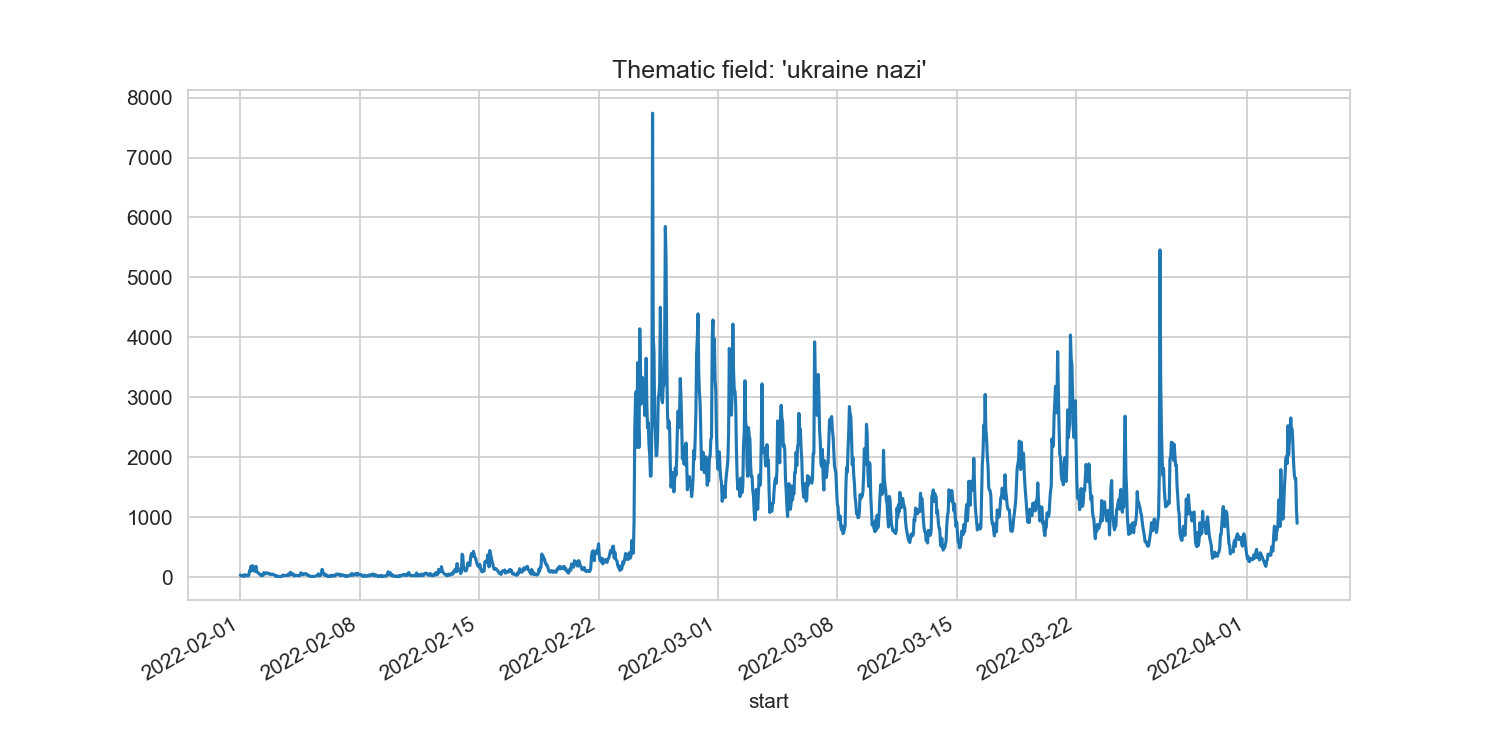}
 \caption{Time series of tweets for the thematic field 'ukraine nazi'}
 \label{ts2}
 \end{figure}

\begin{figure}
\center
 \includegraphics[width=1\linewidth]{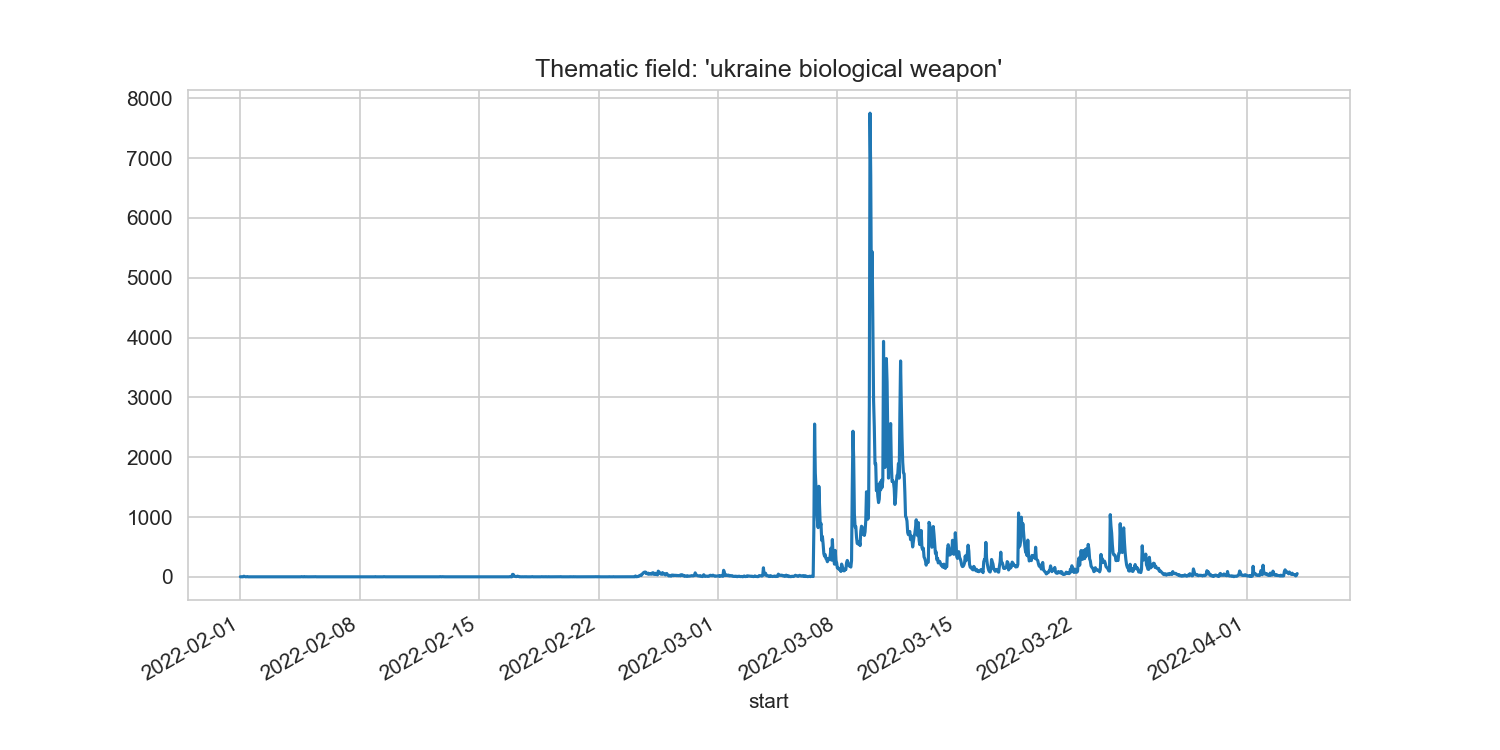}
 \caption{Time series of tweets  for the thematic field 'ukraine biological weapon' }
 \label{ts4}
 \end{figure}

\begin{figure}
\center
 \includegraphics[width=1\linewidth]{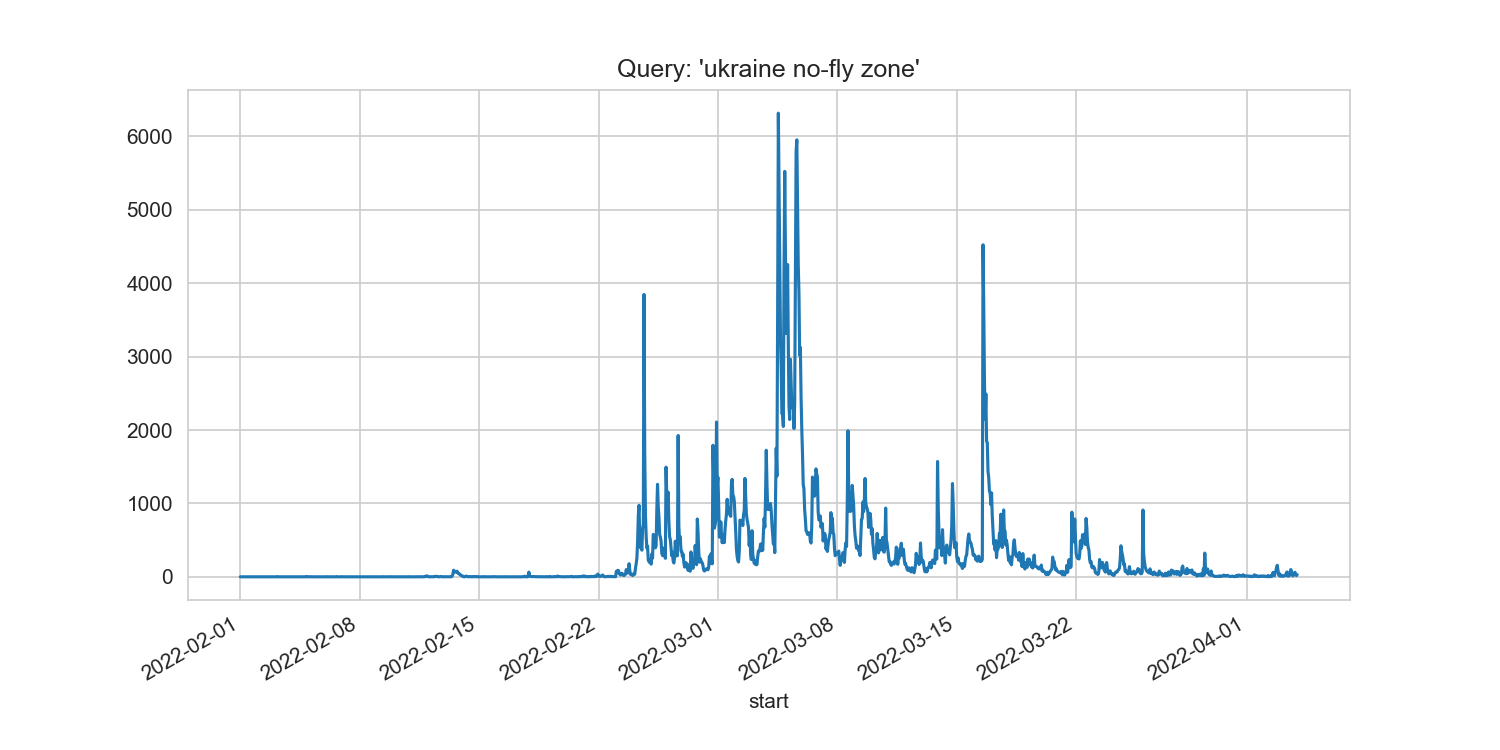}
 \caption{Time series of tweets}
 \label{ts5}
 \end{figure}

For the analysis,  we used a combination of keywords related to 
thematic areas under consideration.  The keywords related to the entity under consideration can be treated as a  thematic field. The use of semantic and thematic fields  for text analytics is considered in~\cite{pavlyshenko2013classification, pavlyshenko2014clustering,
pavlyshenko2014genetic, pavlyshenko2021forming}.  To load tweets, we have used Twitter API v2.
For the 'ukraine nazi' thematic field  the following Twitter API query "(ukraine OR ukrainian OR ukraine's) (nazi OR nazism, nazists OR neonazi OR neo-nazi)" has been used. 
For the 'ukraine biological weapon' thematic field,  the query "ukraine biological (weapon OR weapons OR warfare)" has been used.   For the analysis, the following main python packages were used: \textit{'pandas', 'matplotlib', 'seaborn', 'tweepy'}. Figures~\ref{ts1}-\ref{ts5} show the time series for tweet counts for different queries.  The lower case of query keywords allows searching tweets with  the keywords with both lower and upper cases. 
As the results show,  for the 'ukraine nazi' thematic field, the discussion  of underlying problems   rose dramatically after February 24,  the date of Russian invasion of Ukraine. 
The amount of tweets related to this theme  before that date was at the minimum level. That itself leads to the conclusion that the problem with nazi in Ukraine was just a formal reason to justify the invasion.  Another  claim of Russian propaganda was about biological weapons that were allegedly being developed in Ukrainian laboratories (Figure~\ref{ts4}).
For instance,  it was claimed that a special virus was being developed and it was meant to be distributed through bats and migratory birds.  
 
\section{Deep Learning Approach for Fake News Detection}
\begin{figure}
\center
 \includegraphics[width=1\linewidth]{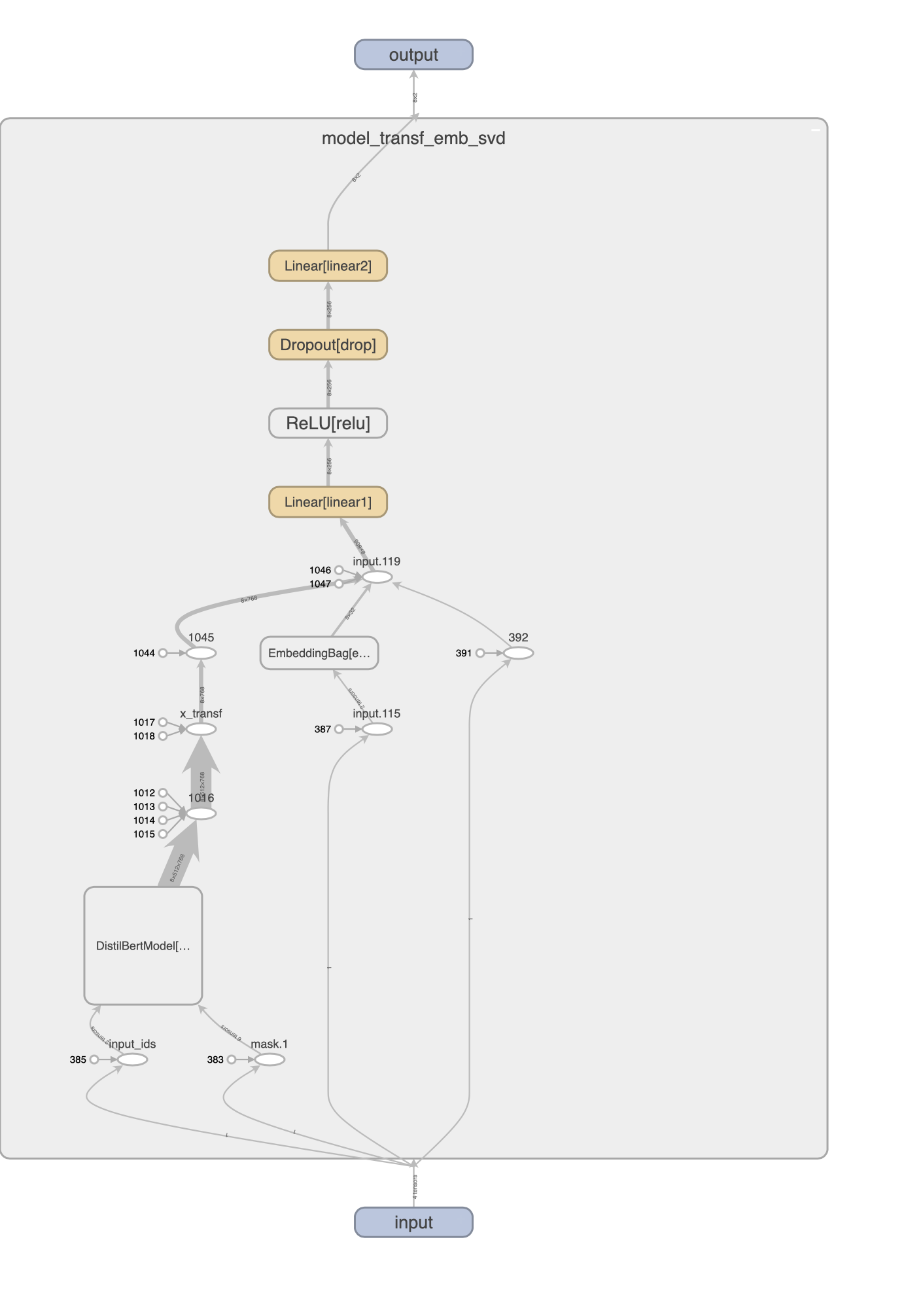}
 \caption{Deep learning model structure}
 \label{fnmodel}
 \end{figure}

\begin{figure}
\center
 \includegraphics[width=0.75\linewidth]{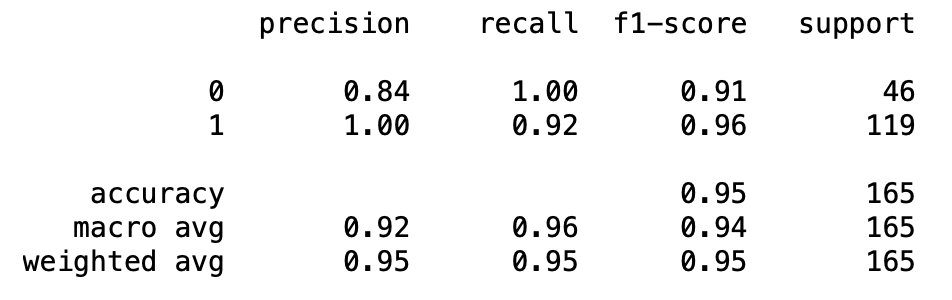}
 \caption{Model evaluation results on the validation dataset}
 \label{fnmodel_evaluation}
 \end{figure}

Fake news can be detected and revealed  by analyzing  facts and comparing them with reality and other news sources.  Let us consider the methods which make it possible to detect fake news automatically.  It is not easy to develop an AI system  which can analyze the facts.  But for manipulative news, it is typical  to  amplify them artificially in different ways,  e.g. by retweeting manipulative tweets  many times using different users' accounts. Some accounts can be  bots which were artificially created, others can belong to real users. It makes it possible to detect fake news using an approach which analyzes the patterns of users' behavior.  Also, fake news have specific patterns in the text of messages. Both users' behavior and text patterns can be captured by deep machine learning algorithms. As the features for a predictive model, we used tweet texts and  the  list of users' usernames  who retweeted those tweets.  For model developing, evaluation and prediction, the Python framework \textit{'pytorch'} was used.  The ML model consists of several concatenated neural subnetworks: subnetwork with DistilBERT transformer which  ingests tokens of tweet text,s  subnetwork with the embedding layer with averaging  which ingests the mixture of encoded words of tweet texts and lists of usernames of retweeters, 
as well as a subnetwork for the components of truncated singular value decomposition of TF-IDF matrix for the list of usernames of retweeters.  Figure~\ref{fnmodel} shows the structure of the deep learning model for fake and manipulative news detection.  For our case study,  the loaded tweets with the thematic fields  'ukraine nazi' and 'ukraine biological weapon' were used. For the model training and evaluation, the  tweet datasets with a specified list of users who retweeted those tweets were  created. For the analysis, only the tweets with a specified threshold for retweet counts were included.  The dataset was labeled using an appropriate tweet id, usernames, hashtags of tweets which can be treated as fake or manipulative.  Figure~\ref{fnmodel_evaluation} shows model evaluation results on the validation dataset. 

The examples of tweets  which were recognized by the deep learning model as fake or manipulation  news for the thematic fields 'ukraine nazi' and 'ukraine biological weapon' are at~\cite{examplesfndet}.  
\section{Detecting Artificially Generated News}
\begin{figure}[H]
\center
 \includegraphics[width=0.75\linewidth]{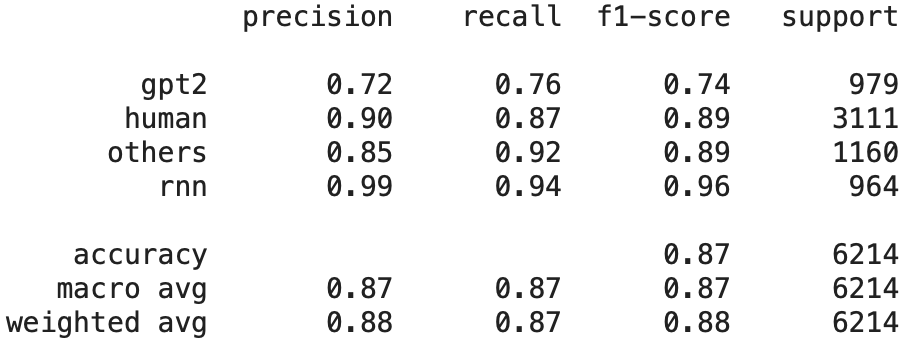}
 \caption{Evaluation results for the model with DisstilBERT transformer layer}
 \label{aitxt1}
 \end{figure}

\begin{figure}[H]
\center
 \includegraphics[width=0.75\linewidth]{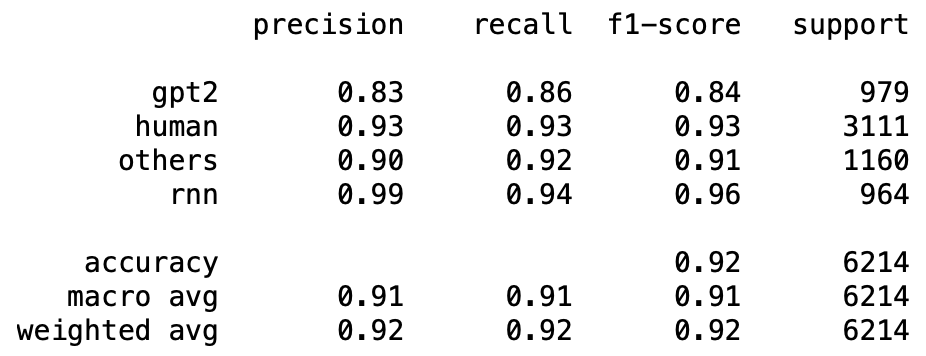}
 \caption{Evaluation results for the model with the concatenation DisstilBERT transformer layer and embedding layer for usernames}
 \label{aitxt2}
 \end{figure}

One of the ways to form manipulative trends is to produce artificially created news tweet messages~\cite{jawahar2020automatic,fagni2021tweepfake, murayama2021dataset}, e.g.  by paraphrasing an initial text using Seq2Seq neural network models.  For this purpose,  GPT-2, and  BART transformers can be used. Each pretrained transformer has its own patterns for generating texts using an encoder-decoder approach.  These patterns can be detected by other transformers which are fine-tuned on a dataset with an artificialy generated text. For the model fine-tuning,  a TweepFake dataset of artificially generated texts~\cite{fagni2021tweepfake} was used.  The framework \textit{'pytorch'} was used for the model developing and evaluation. 
We tried two  models -- the neural network with the DistilBERT transformer layer and the neural network with the concatenation of the DistilBERT transformer with embedding layers for the usernames of users who post tweets.  Figures~\ref{aitxt1}, ~\ref{aitxt2} show the  model evaluation on the validation dataset for these two models. 
 As the results show,  the embedding layer for usernames can improve accuracy scores, it means that usernames  have predictive potential  for detecting artifitially created tweets.  We applied the fine-tuned model to the datasets of  tweets of the semantic fields  under consideration and have received the following results: human - 80\%, GPT-2-10\%, RNN - 0.3\%, Others - 3\%.
\section{The Impact of the Discussion About Company's Behavior On Its Stock Prices}
\begin{figure}
\center
 \includegraphics[width=0.95\linewidth]{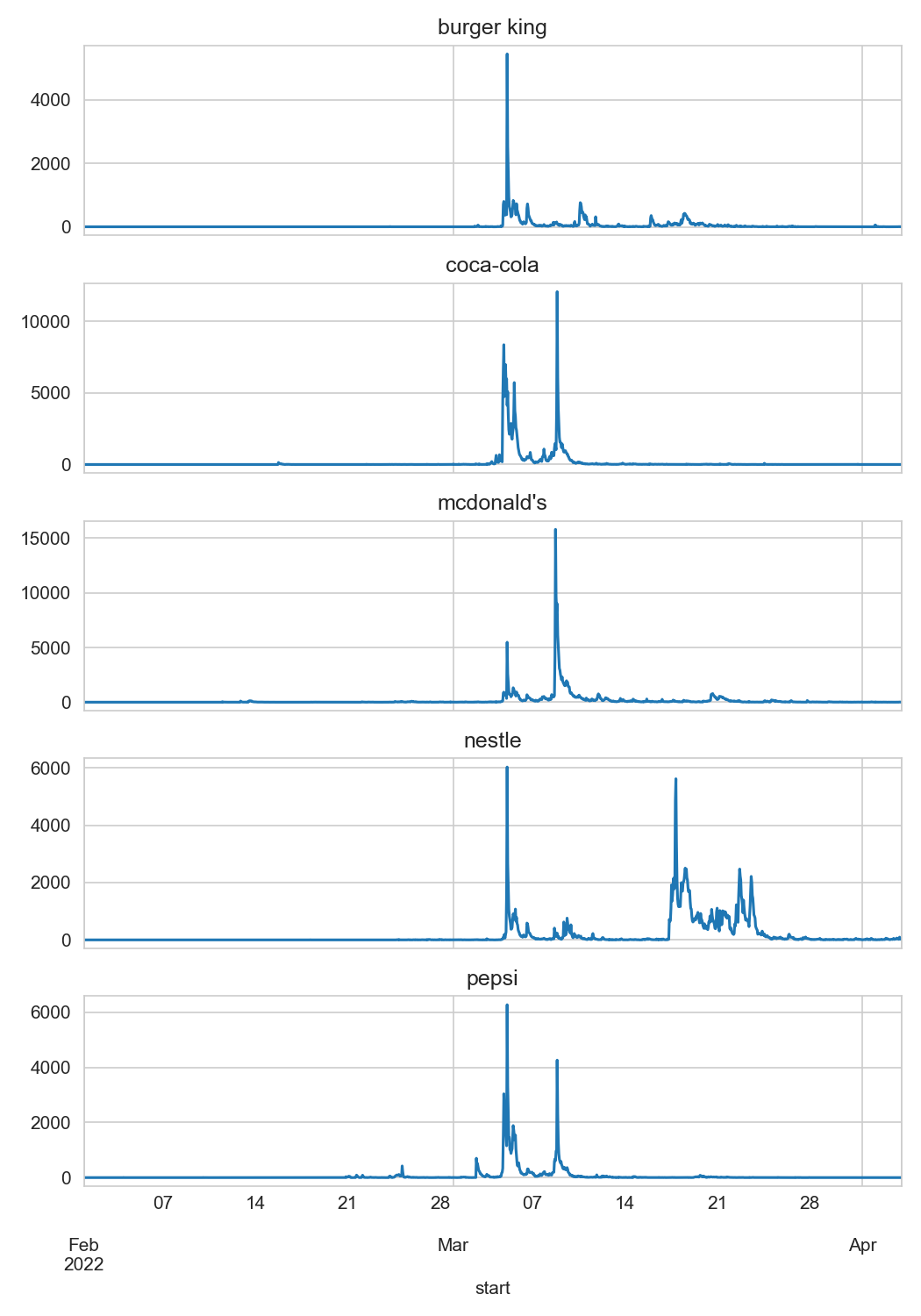}
 \caption{Time series of tweets for  trends related to different companies}
 \label{twtscomp1}
 \end{figure}

\begin{figure}
\center
 \includegraphics[width=0.75\linewidth]{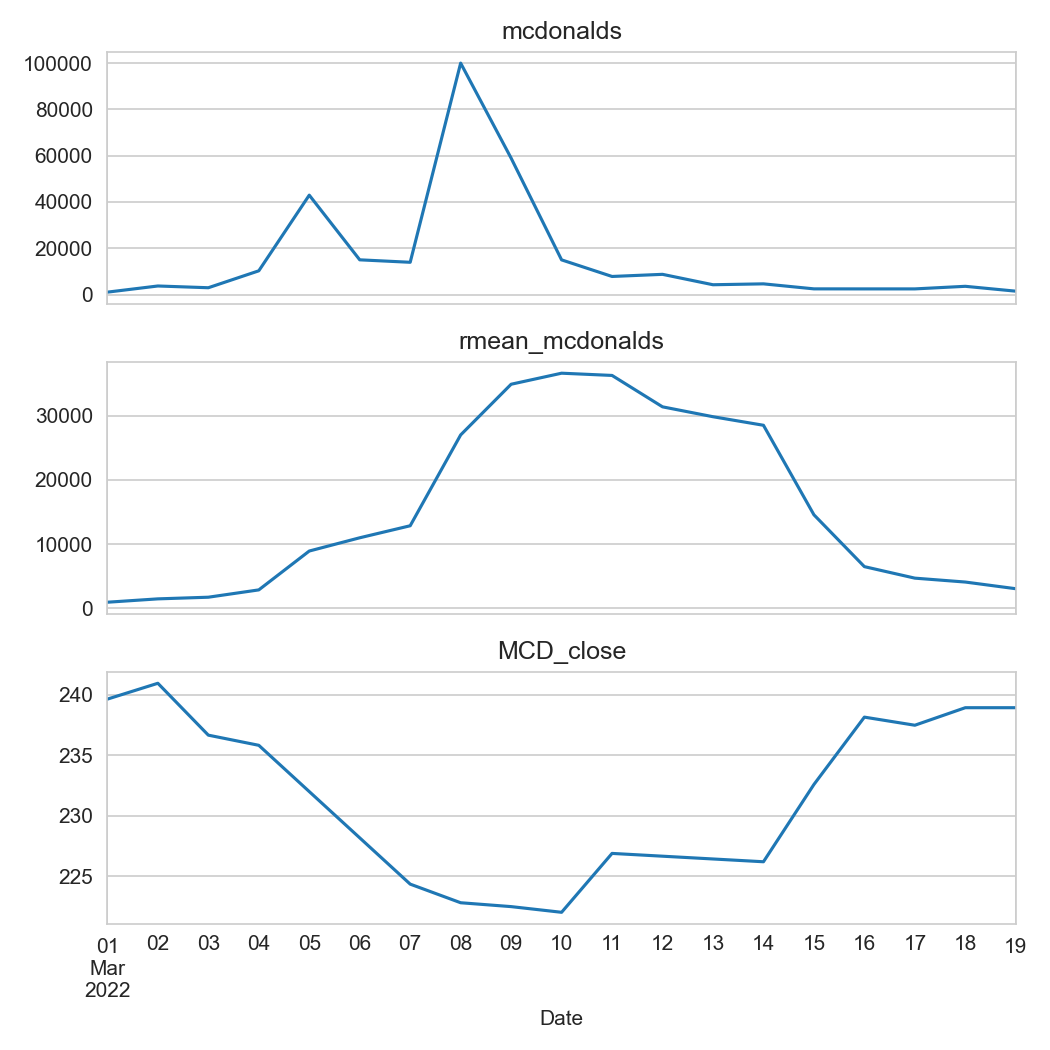}
 \caption{Time series of tweets related to McDonald's,  rolling mean of the time series, stock price time series for the 'MCD' ticker}
 \label{twtsmcd1}
 \end{figure}

\begin{figure}
\center
 \includegraphics[width=0.75\linewidth]{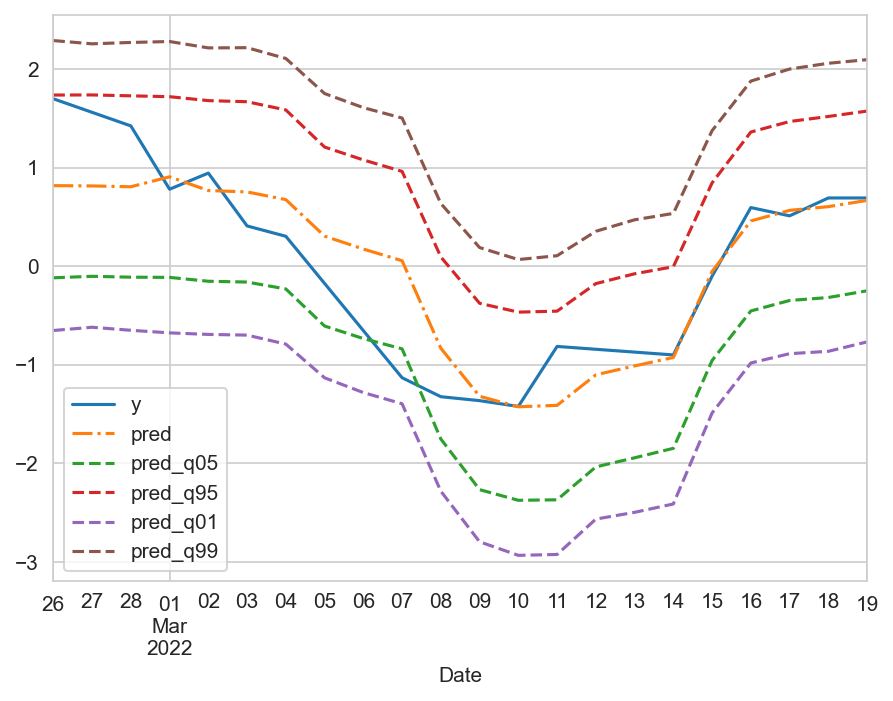}
 \caption{Normalized time series of stock price for the 'MCD' ticker (y), prediction for stock price (pred),  quantiles for prediction (0.01 - pred\_q01, 0.05 - pred\_q05, 0.95 - pred\_q95,  0.99 - pred\_q99 )}
 \label{stockpricemcd}
 \end{figure}

\begin{figure}
\center
 \includegraphics[width=0.75\linewidth]{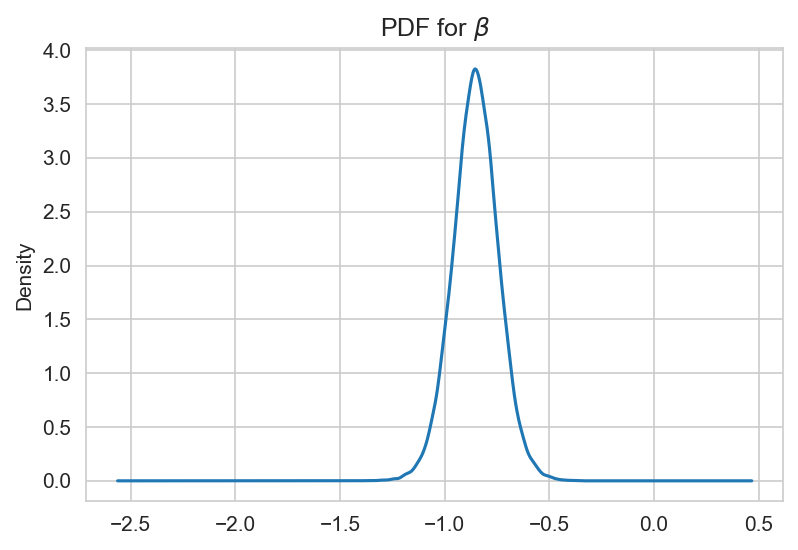}
 \caption{Probability density function for  $\beta$ model parameter}
 \label{pdfbetamcd}
 \end{figure}

Russian invasion of Ukraine has huge impact on users' attitude to different companies and shapes discussion trends on Twitter. 
Figure~\ref{twtscomp1} shows time series for tweet counts related to the consideration of different companies in Russia during Russian invasion of Ukraine.  For the case study, let us consider  McDonald's.  Figure~\ref{twtsmcd1} shows the tweet counts in the trend of considering this company,  a rolling mean of this time series, with a 7-day window and stock price for the 'MCD'  ticker for the McDonald's company. 
As a result of public discussion,  it is known that McDonald's has stopped its activity in Russia.  After this announcement,  the stock price for the 'MCD' ticker  has returned to almost initial price value before the invasion  period (Figure~\ref{twtsmcd1}). The  stock price time series for the 'MCD' ticker  were loaded using Python package \textit{'yfinance'}.
It shows that public consideration of a company's behavior which is reflected on the social networks can have impact on the stock market.  
It is supposed that consideration has some cumulative effect, that is why the rolling mean of the time series of tweet counts corresponds more precisely to company's stock prices on the stock market. 
Let us analyze the   dependency  of stock prices  on the tweet counts rolling mean.  It is important to estimate the uncertainty of such dependency.  For the modeling of such a dependency, the  Bayesian regression was applied. The probabilistic model can be considered as follows:
\begin{equation}
\begin{split}
&y \sim  Student_t (\nu, \mu, \sigma),\\
&\mu = \alpha + \beta x, 
\label{eq_1}
\end{split}
\end{equation}
where $\nu$ is a distribution parameter,  known as degrees of freedom.
Target variable $y$ is described by Student's t-distribution 
which has fat tails that makes it possible to take into account extreme events and values that enable us  to estimate uncertainties more accurately. 
 For Bayesian inference calculations, we used a  Python package \textit{'pystan'}  for Stan platform for statistical modeling \cite{carpenter2017stan}. For the  analysis, as a feature independent variable,  we used z-scores for the rolling mean of tweet counts time series, as a target variable, we used z-scores for stock price for  'MCD' ticker.  As a result of sampling,  the mean value as well as  quantiles 0.01, 0.05, 0.95, 0.99  for the target variable were calculated. 
 Figure~\ref{stockpricemcd}  shows the results of modeling.  
 Figure~\ref{pdfbetamcd} shows the probability density function for $\beta$ parameter. 
 Quantile 0.05  for predicted target variable can be treated as the value of risk (VaR) that is a quantitative characteristic  for risk assessment.
 
\section{Analysis of Tweets Using Frequent Itemsets}
\begin{figure}
\center
 \includegraphics[width=0.75\linewidth]{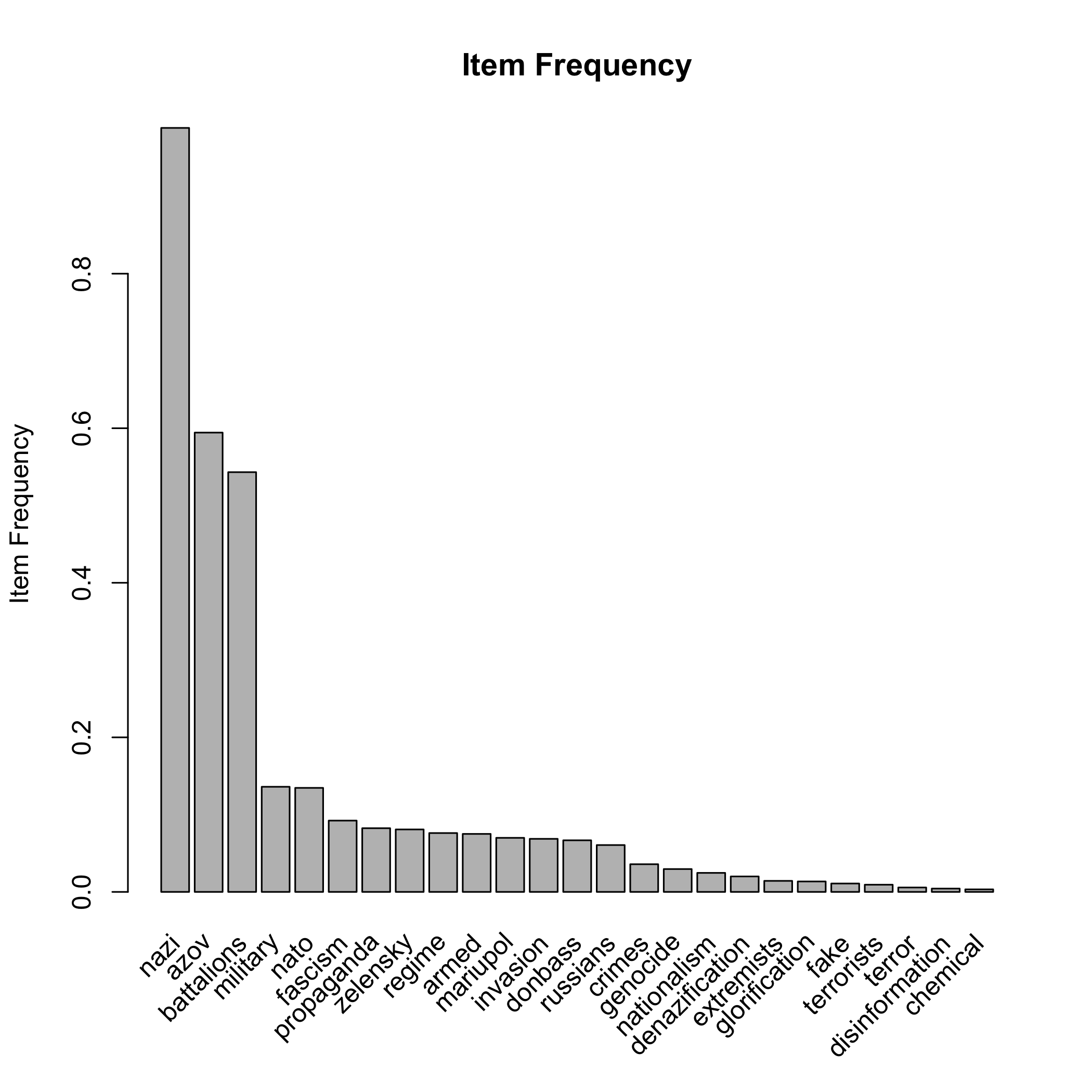}
 \caption{Keyword  frequencies related to the thematic field 'ukraine nazi'}
 \label{freq_1}
 \end{figure}

\begin{figure}
\center
 \includegraphics[width=1\linewidth]{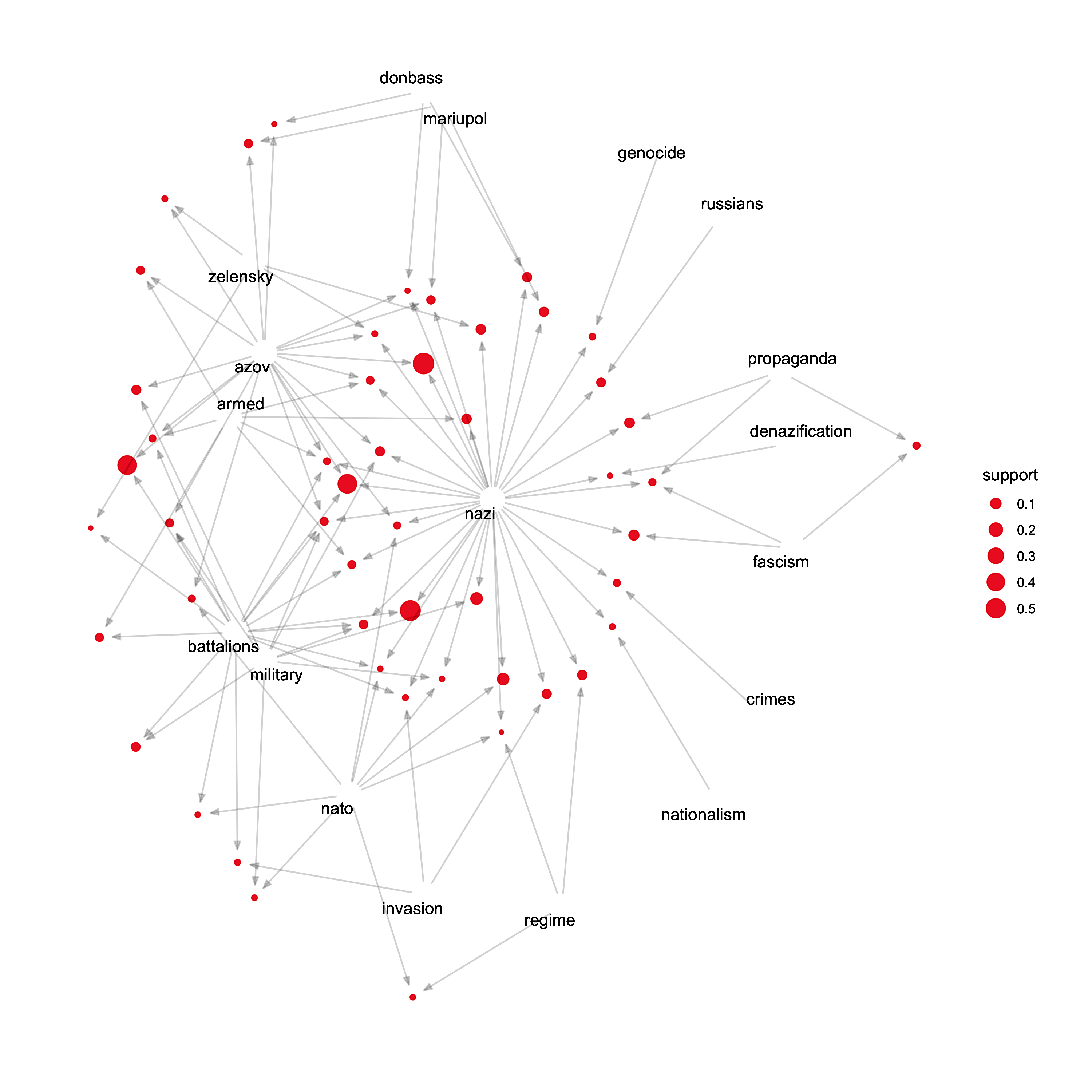}
 \caption{Graph of  semantic frequent itemsets}
 \label{fritemsets_nazi_1}
 \end{figure}

\begin{figure}
\center
 \includegraphics[width=0.75\linewidth]{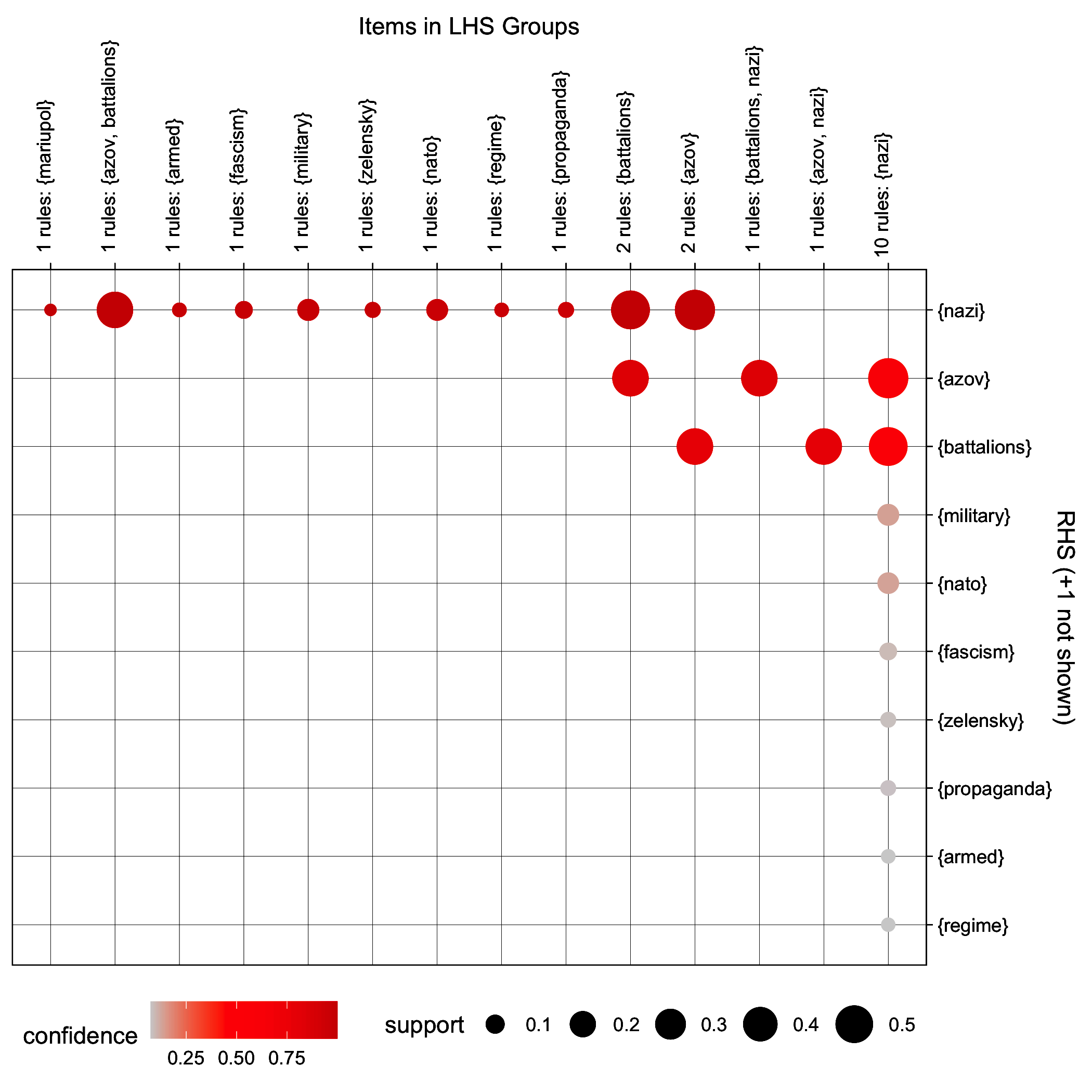}
 \caption{Association rules   represented by a grouped matrix}
 \label{rules_nazi_grouped}
 \end{figure}

\begin{figure}
\center
 \includegraphics[width=0.75\linewidth]{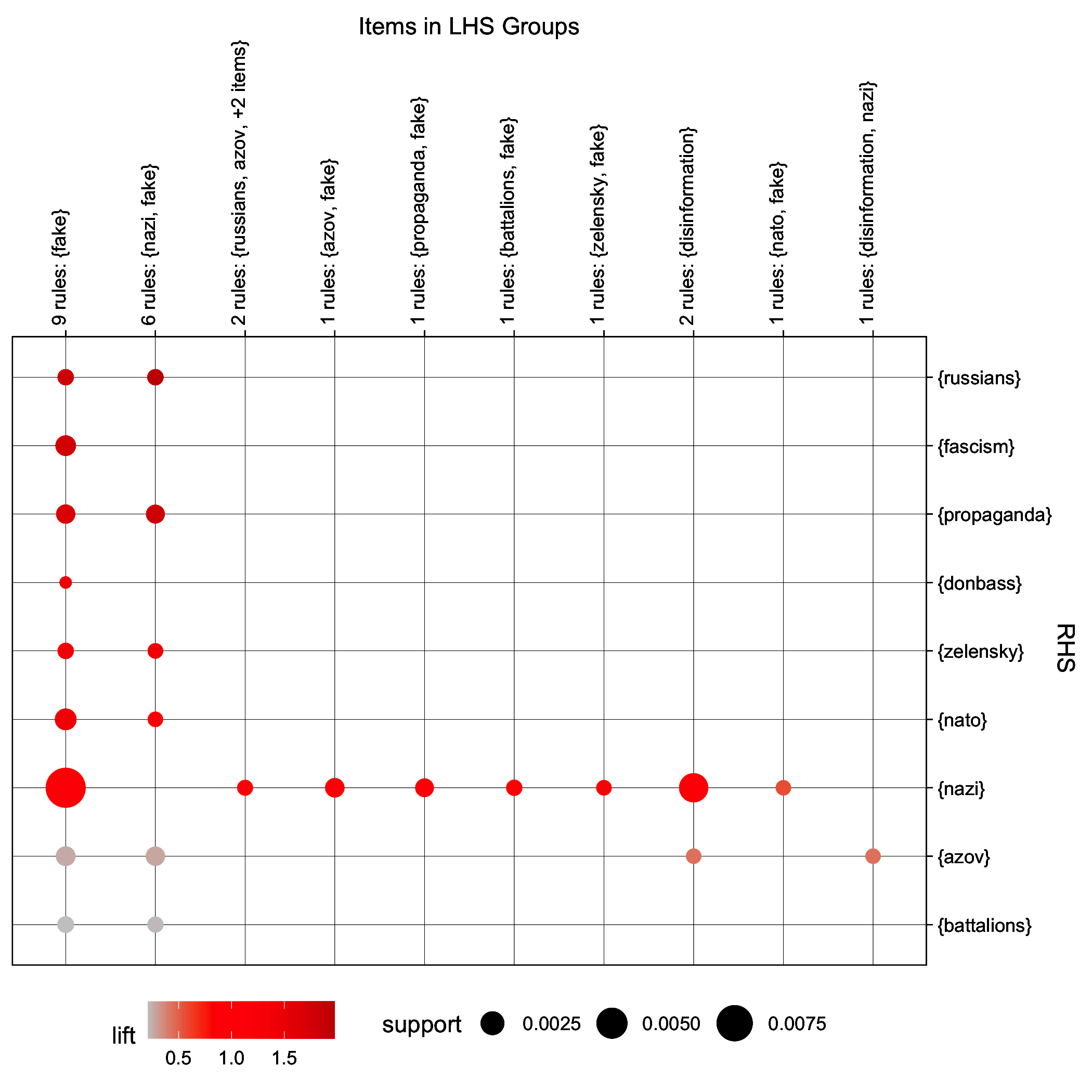}
 \caption{Association rules   represented by a grouped matrix with the keyword 'fake'}
 \label{rules_nazi_fake_grouped}
 \end{figure}

\begin{figure}
\center
 \includegraphics[width=0.75\linewidth]{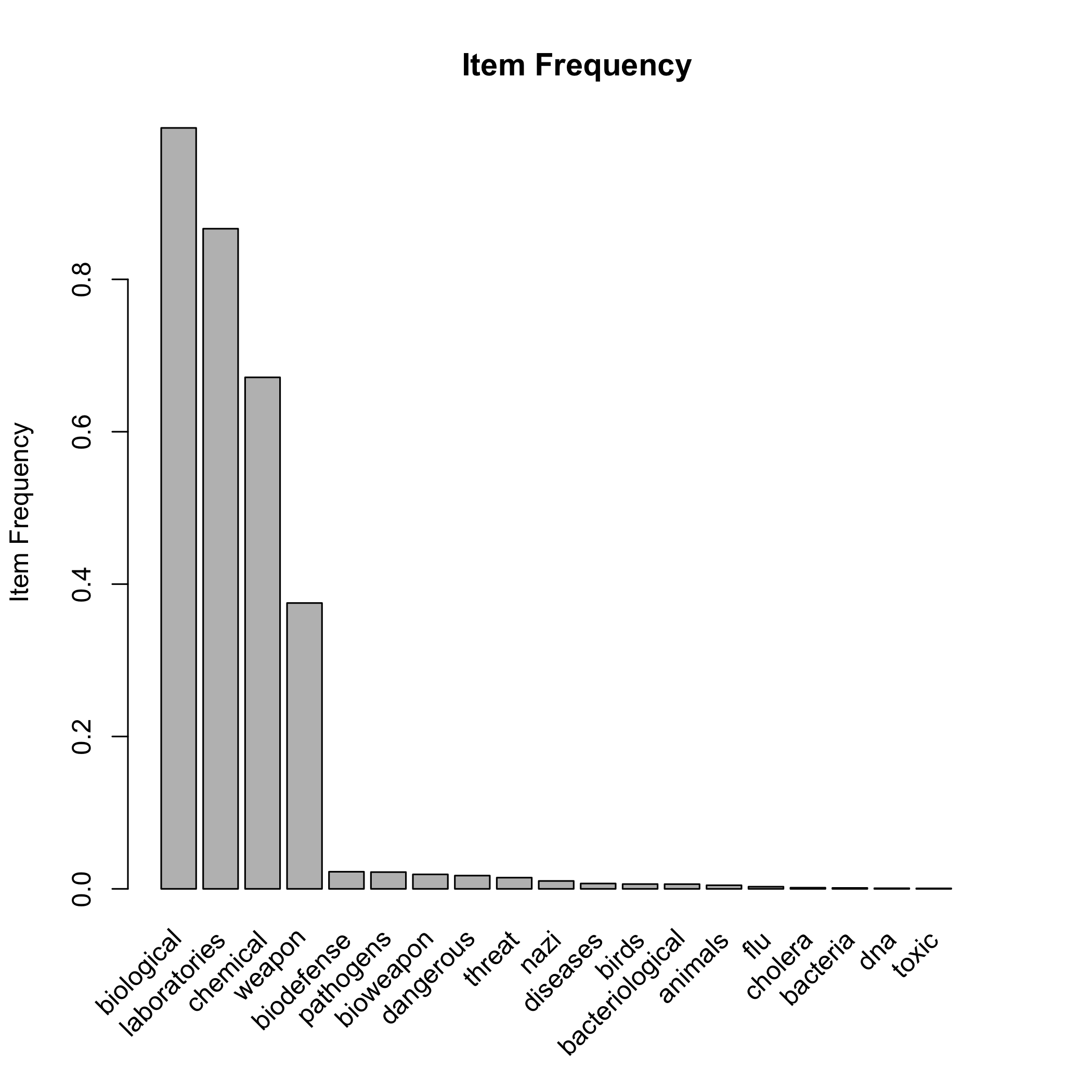}
 \caption{Keyword  frequencies related to the thematic field 'ukraine biological weapon'}
 \label{freq_bioweopns}
 \end{figure}

\begin{figure}
\center
 \includegraphics[width=1\linewidth]{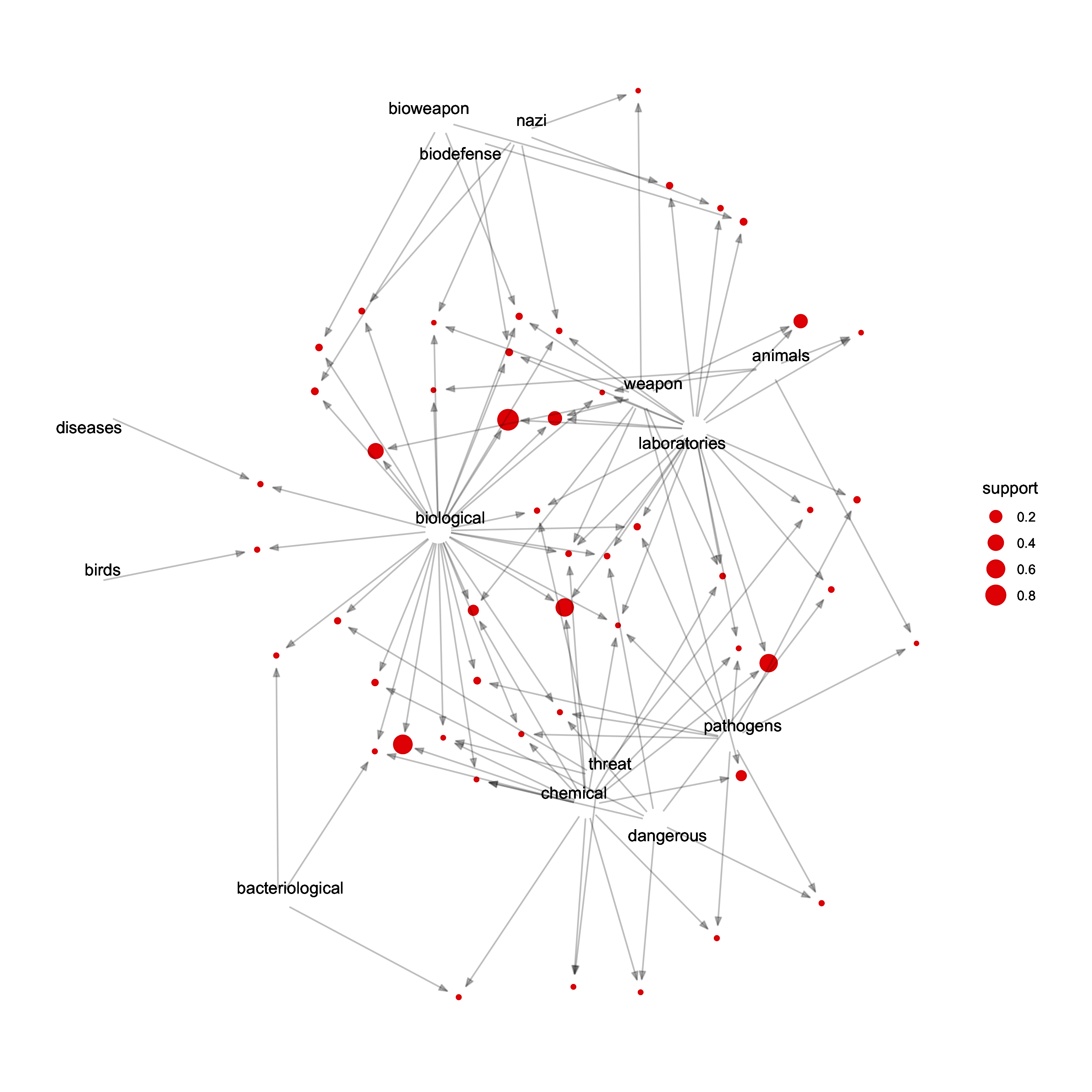}
 \caption{Graph of  semantic frequent itemsets}
 \label{fritemsets_bioweapon_1}
 \end{figure}

\begin{figure}
\center
 \includegraphics[width=0.75\linewidth]{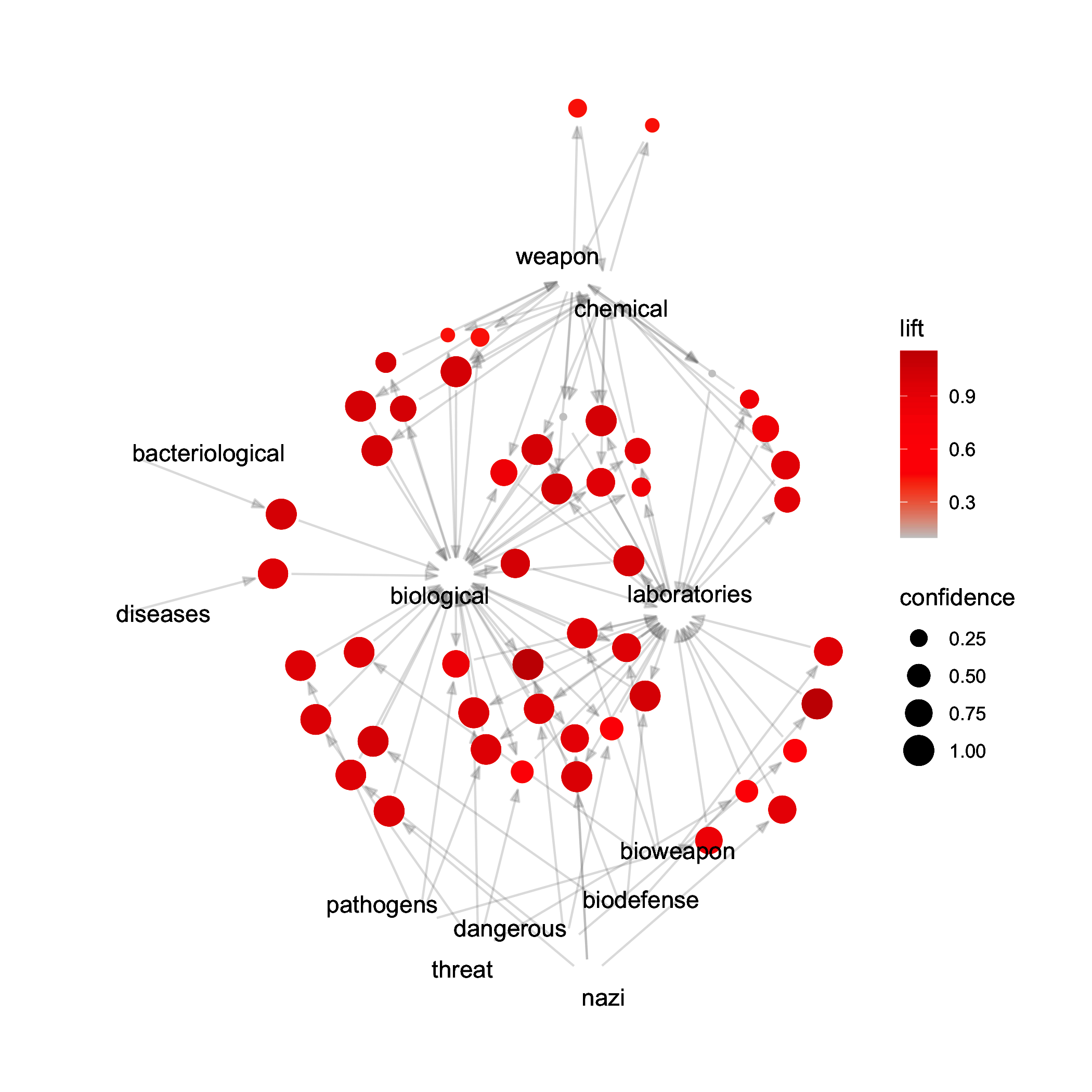}
 \caption{Graph of association rules}
 \label{rules_bioweapon_1}
 \end{figure}

\begin{figure}
\center
 \includegraphics[width=0.75\linewidth]{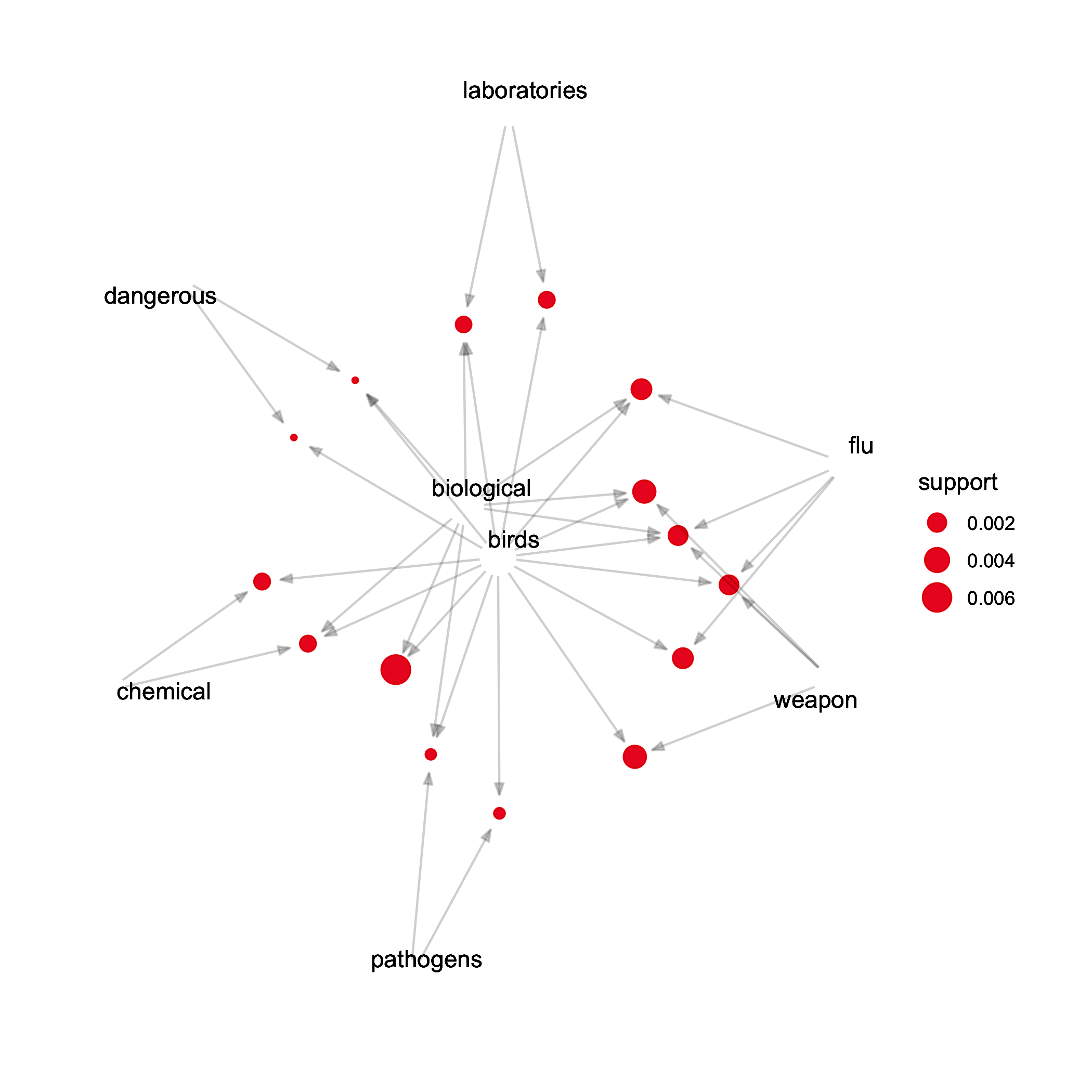}
 \caption{Graph of  semantic frequent itemsets with the keyword 'bird'}
 \label{fritemsets_bioweapon_bird}
 \end{figure}

\begin{figure}
\center
 \includegraphics[width=0.75\linewidth]{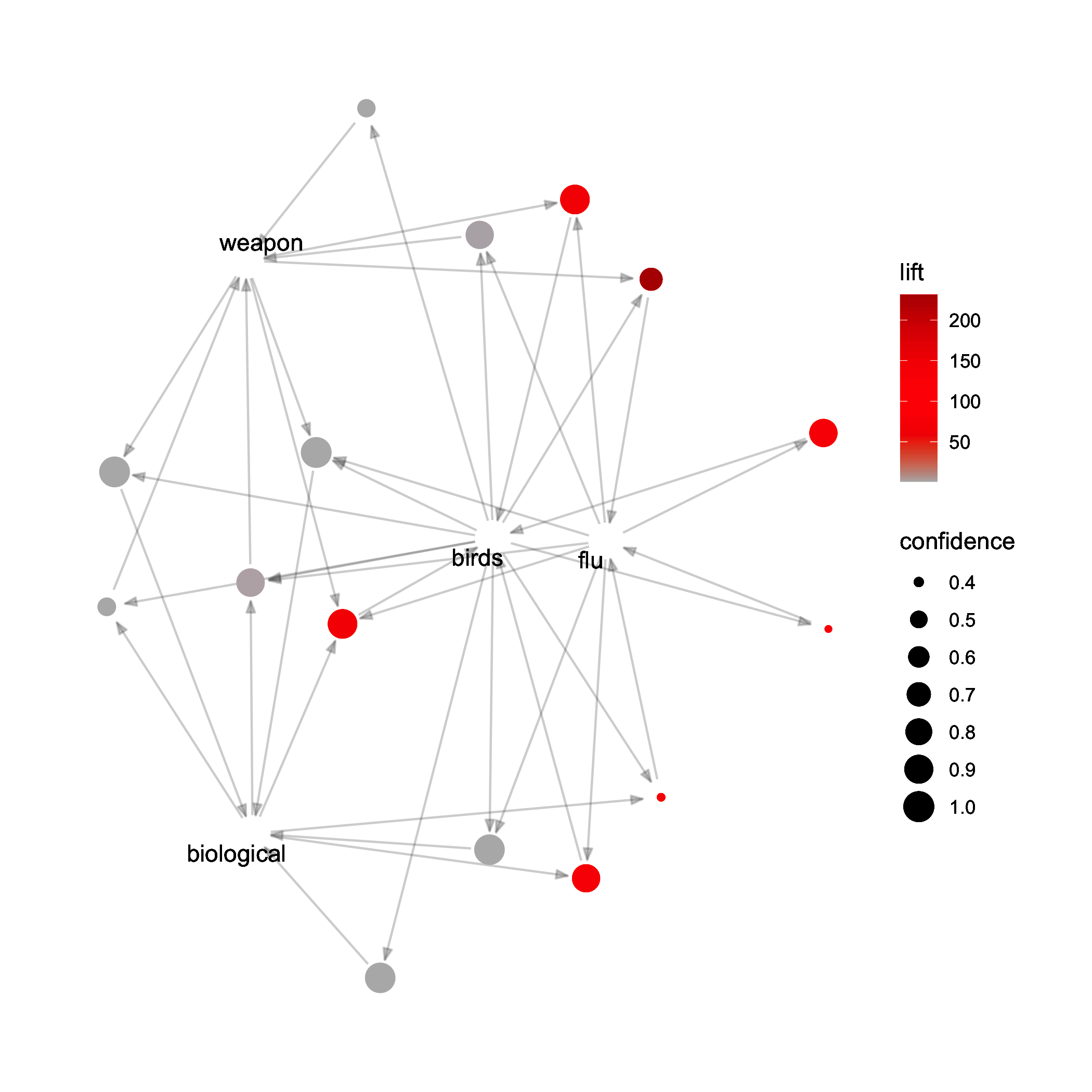}
 \caption{Graph of association rules with the keyword 'bird'}
 \label{rules_bioweapon_bird}
 \end{figure}

\begin{figure}
\center
 \includegraphics[width=0.75\linewidth]{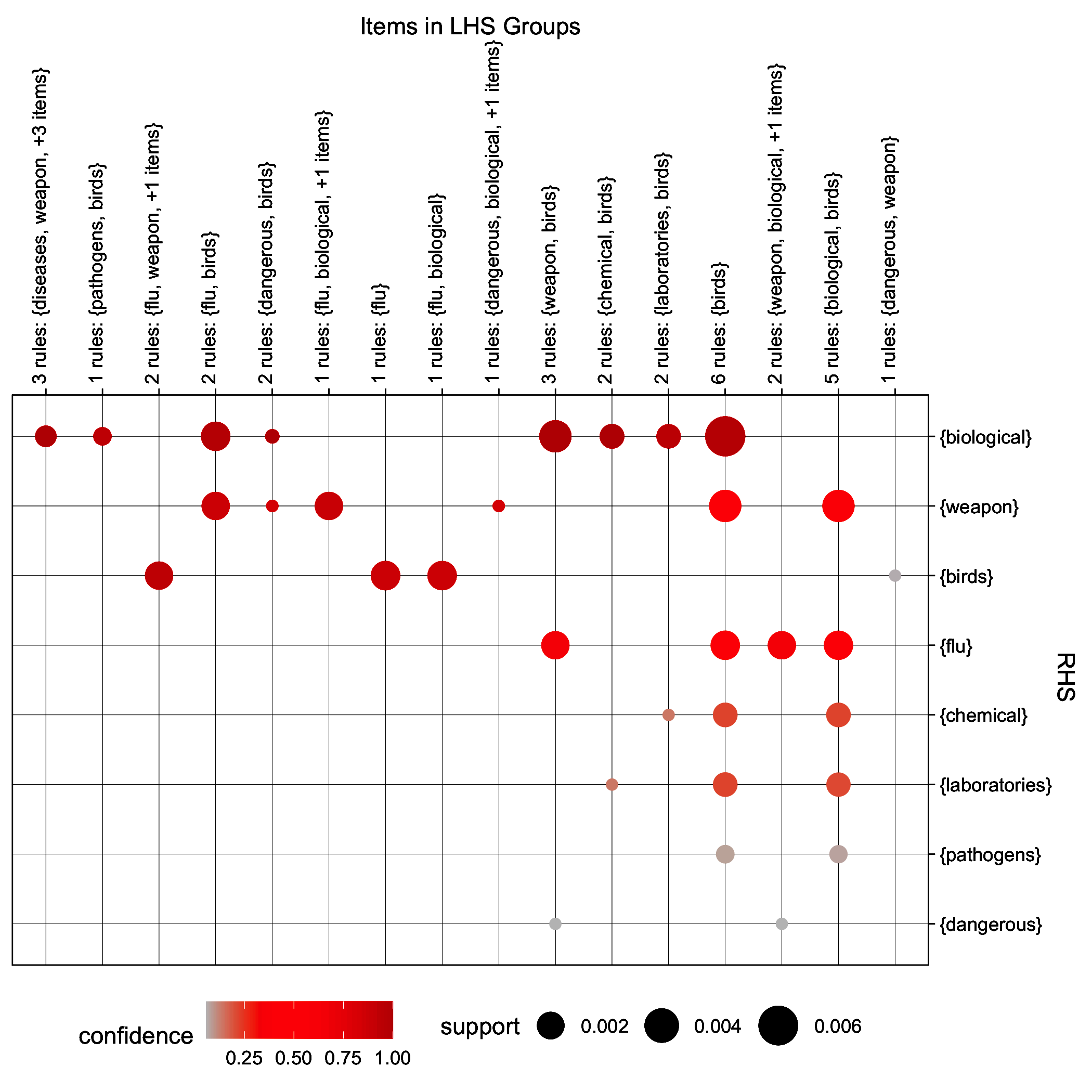}
 \caption{Association rules with the keyword 'bird' represented by a grouped matrix}
 \label{rules_bioweapon_bird_grouped}
 \end{figure}

The frequent itemsets  and associative rules theory is often used in the intellectual analysis~\cite{agrawal1994fast,agrawal1996fast,chui2007mining,gouda2001efficiently,
srikant1997mining,klemettinen1994finding,pasquier1999discovering,
brin1997beyond}.
 It can be used in a text data analysis to identify and analyze certain sets of objects, which are often found in large arrays and are characterized by certain features. 
Let us consider the algorithms for detecting frequent sets and associative rules on the example of processing microblog messages on tweets.  
Figure~\ref{freq_1} shows keyword  frequencies for  the specified thematic field 'ukraine nazi'.
Using these keywords, one can calculate frequent itemsets. 
Figure~\ref{fritemsets_nazi_1} shows the graph of frequent itemsets which describes the semantic structure of entities for a specified thematic field. 
Figure~\ref{fritemsets_nazi_1} shows the graph of the subset of association rules,  Figure~\ref{rules_nazi_grouped} shows the  association rules   represented by a grouped matrix. Figure~\ref{rules_nazi_fake_grouped} shows the association rules which contain the  keyword 'fake'.
Figures~\ref{freq_bioweopns}--\ref{rules_bioweapon_1} show the similar calculation for the thematic field 'ukraine biological weapon'.  Figures~\ref{fritemsets_bioweapon_bird}--\ref{rules_bioweapon_bird_grouped} show the subset of these frequent itemsets and association rules which contain the keyword 'bird'.  The quantitative characteristics of frequent itemsets  like value of support can be used as a predictive feature in machine learning models. 
   \begin{figure}
\center
 \includegraphics[width=0.75\linewidth]{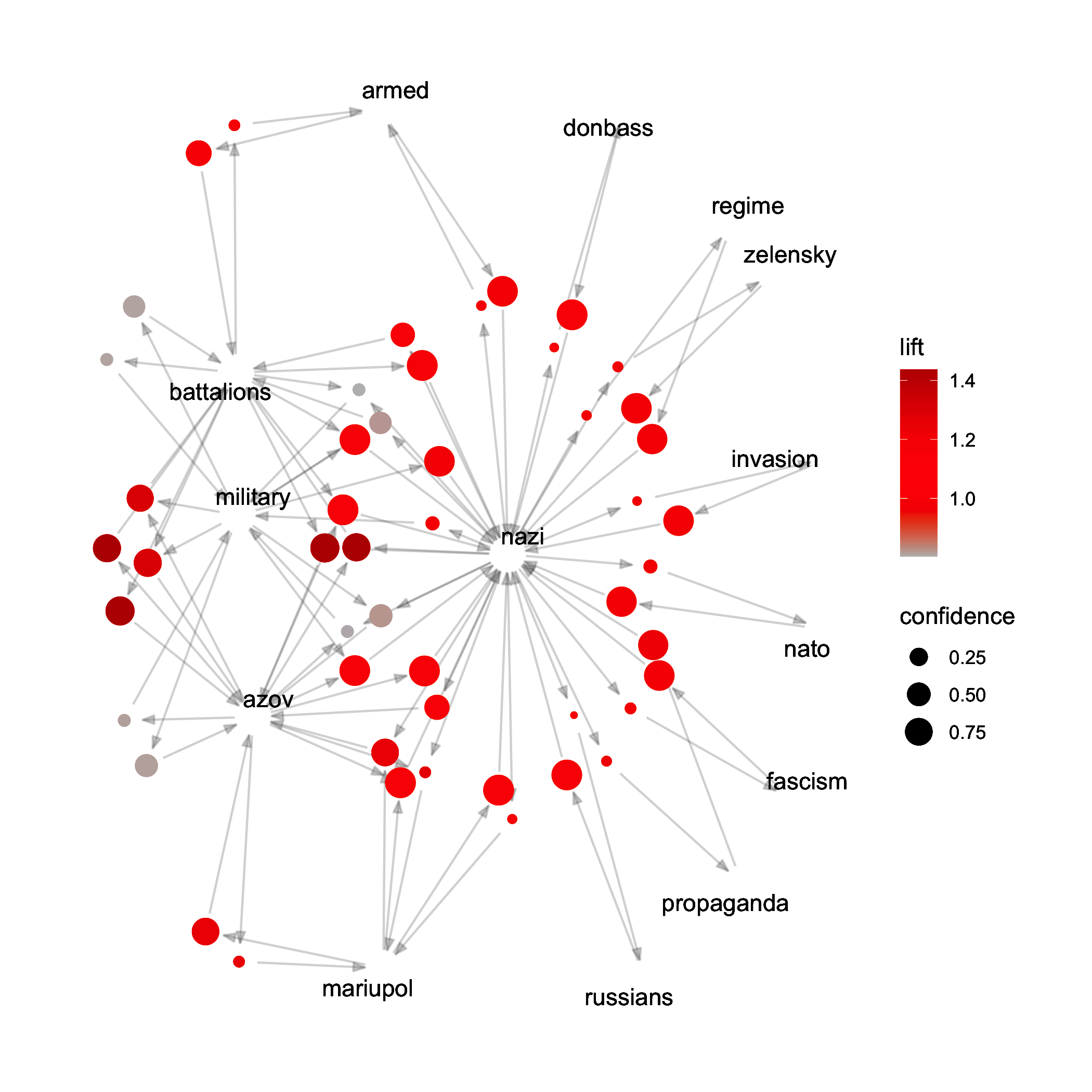}
 \caption{Graph of association rules}
 \label{/rules_nazi_1}
 \end{figure}
 
 \section{Graph Structure of Tweets}
\begin{figure}
\center
 \includegraphics[width=0.85\linewidth]{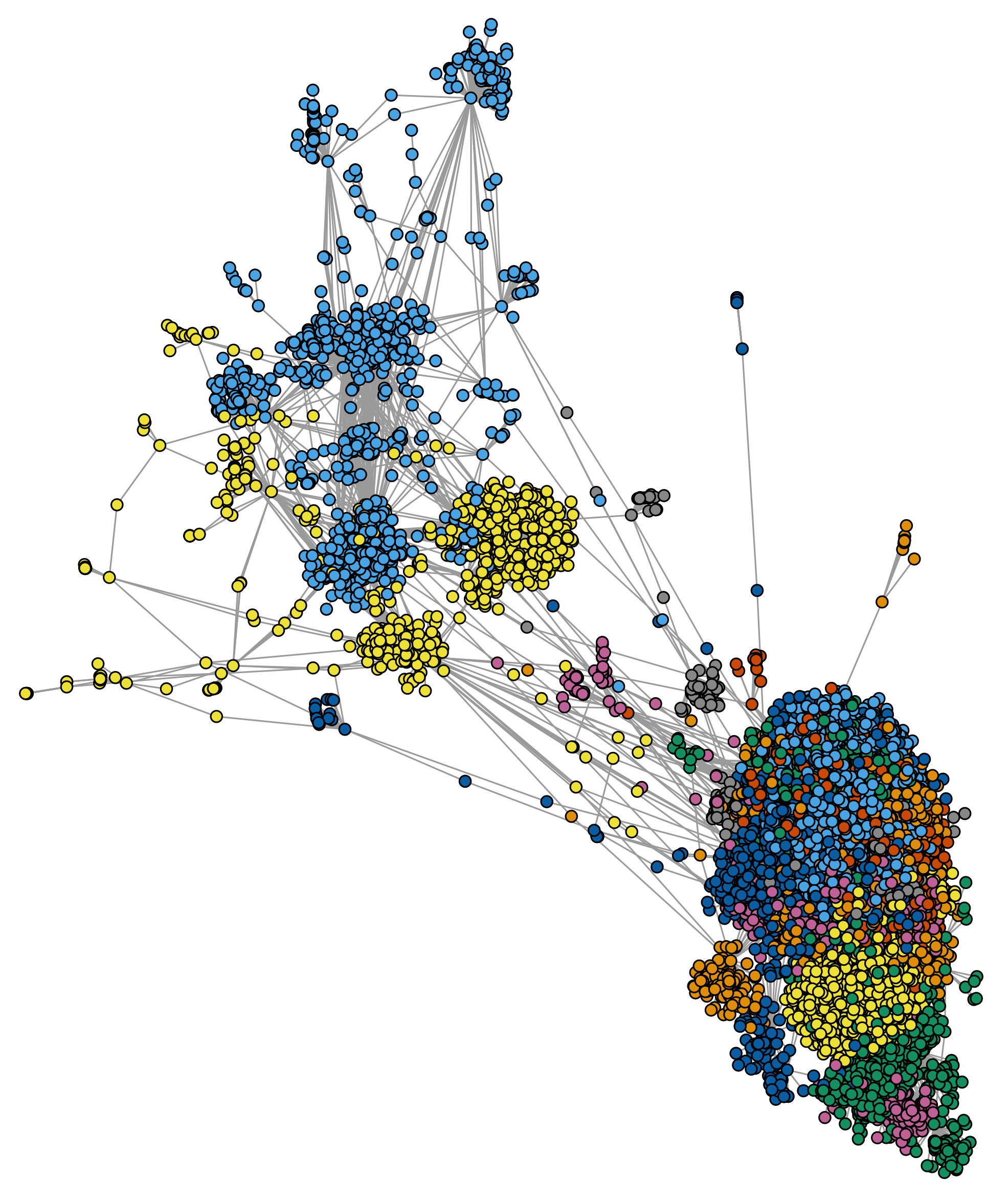}
 \caption{Graph of users' connections}
 \label{usr_graph}
 \end{figure}

The  relationships among users can be considered as a graph, where vertices denote users and edges denote their connections.
Using graph mining algorithms, one can detect user communities and find ordered lists of users by various characteristics, such as
\textit {Hub, Authority, PageRank, Betweenness}. To identify  user communities, we used the \textit{Community Walktrap} algorithm and  to visualize them we used  \textit{Fruchterman-Reingold} algorithm, which are implemented in the package
 \textit{'igraph'}~\cite{csardi2006igraph} for the \textit{R} programming language environment. 
The \textit{Community Walktrap} algorithm searches for related subgraphs, also called communities, by random walk~\cite{pons2005computing}.
A graph which shows the relationships between users can be represented by
 Fruchterman-Reingold algorithm~\cite{fruchterman1991graph}.
 The qualitative structure of user's connections  can be used for aggregating different quantitative time series and, in such a way, creating new features for predictive models which can be used, for example, for predicting target variables.  
 Figure~\ref{usr_graph} shows users connections and revealed communities for the subset of tweets which are related to the trends under consideration.  
 The results show that some communities marked by different colors are highly isolated and have only few connections outside.  This kind of communities can be treated as suspicious,  since artificially created communities for amplifying manipulative news are also highly isolated and their activity is often concentrated on amplifying by retweeting tweets from a limited set of users. Therefore,  the numerical characteristics of users communities can have a predictive potential. 
\section{Conclusion}
The obtained results show that 
an effective system for detecting fake and manipulative news can be developed 
 using combined neural network which consist of three concatenated subnetworks: the subnetwork with DistilBERT transformer for tweet texts, the subnetwork with embedding of tweet words and  usernames of users who retweeted tweets,  and the  subnetwork for the components of singular value decomposition of TF-IDF matrix for lists of usernames of users who retweeted tweets.  Discussions on social networks about companies behavior has impact on their business and their stock prices on the stock market. 
To analyze such an impact and make risk assessment, Bayesian regression can be used. 
Using the theory of frequent itemsets and association rules along with thematic fields of keywords, makes it possible to reveal the semantic structure for entities in news messages. The quantitative characteristics association rules like support and confidence can be used as features for machine learning predictive models.  Using the graph theory, the hidden communities of users can be revealed and their characteristics can be used in a machine learning approach for fake and manipulative news detection.

\bibliographystyle{ieeetr}
\bibliography{article.bib}
\end{document}